\documentclass[letterpaper]{article} 
\usepackage[preprint]{aaai2027}
\usepackage[hyphens]{url}  
\usepackage{graphicx} 
\usepackage{natbib}  
\usepackage{caption} 
\usepackage{booktabs}
\usepackage{amsmath}
\title{Paint What You See: Benchmarking Dexterous Visual Tool Use in Multimodal Agents}
\author{
    Shudong Liu\equalcontrib\textsuperscript{\rm 1,\rm 3},
    Dongyang Chen\equalcontrib\textsuperscript{\rm 2,\rm 3},
    Enci Zhang\textsuperscript{\rm 1},
    Jinwei Liang\textsuperscript{\rm 3},
    Zheng Ma\textsuperscript{\rm 3},
    Lewei Lu\textsuperscript{\rm 3}\corresponding
}

\newcommand{\iconlink}[2]{%
  \raisebox{-0.12em}{\includegraphics[height=0.95em]{#1}}\,\url{#2}%
}

\affiliations{
    \textsuperscript{\rm 1}Peking University,\\
    \textsuperscript{\rm 2}Tsinghua University,\\
    \textsuperscript{\rm 3}SenseTime Research\\
    \{shooo, eczhang\}@stu.pku.edu.cn,
    chen-dy25@mails.tsinghua.edu.cn,
    \{liangjinwei, mazheng2, luotto\}@sensetime.com\\[4pt]
    \iconlink{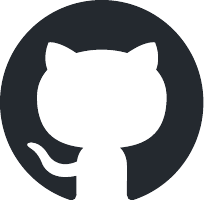}{https://github.com/OOOHS/EASEL}
    \quad
    \iconlink{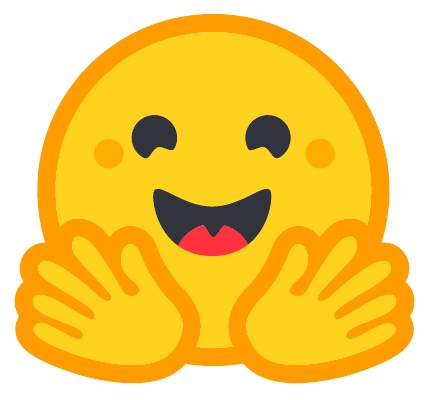}{https://huggingface.co/EASEL-Bench}
}
\begin{document}

\maketitle

\begin{abstract}
Evaluation is shifting from static QA toward agentic settings where models act through external tools. We identify a critical yet underexplored capability within this space—dexterous visual tool use: fine-grained, closed-loop parameterized visual action in which models infer tool parameters from visual evidence, and those parameters directly govern the final result. Existing benchmarks cover web navigation, GUI operation, and software engineering, but rarely target this coupling between visual evidence and execution precision.
We propose EASEL, a benchmark evaluating a controlled instance of dexterous visual tool use that adopts reference-guided visual reconstruction as its primary proxy task: the agent incrementally paints a canvas to match a reference image. EASEL additionally includes semantic tasks spanning region annotation, handwriting, and path planning. We further provide EASEL-Data, a 440k-sample two-stage curriculum dataset for trajectory supervision, and EASEL-9B to investigate its effect on this capability.
Evaluation of 25 models reveals that current multimodal agents systematically struggle on EASEL. Reconstruction similarity bottlenecks at low levels (0.40–0.54), while trajectory diagnostics expose severe closed-loop instability—models typically saturate early or degrade post-peak. Semantic tasks reveal sharp capability boundaries in precision annotation and path planning. EASEL-9B, trained on EASEL-Data, surpasses the base model by a relative 6.3\%, ranking third among all evaluated models.
\end{abstract}

\section{Introduction}

\begin{figure}[t]
\centering
\includegraphics[width=0.9\columnwidth]{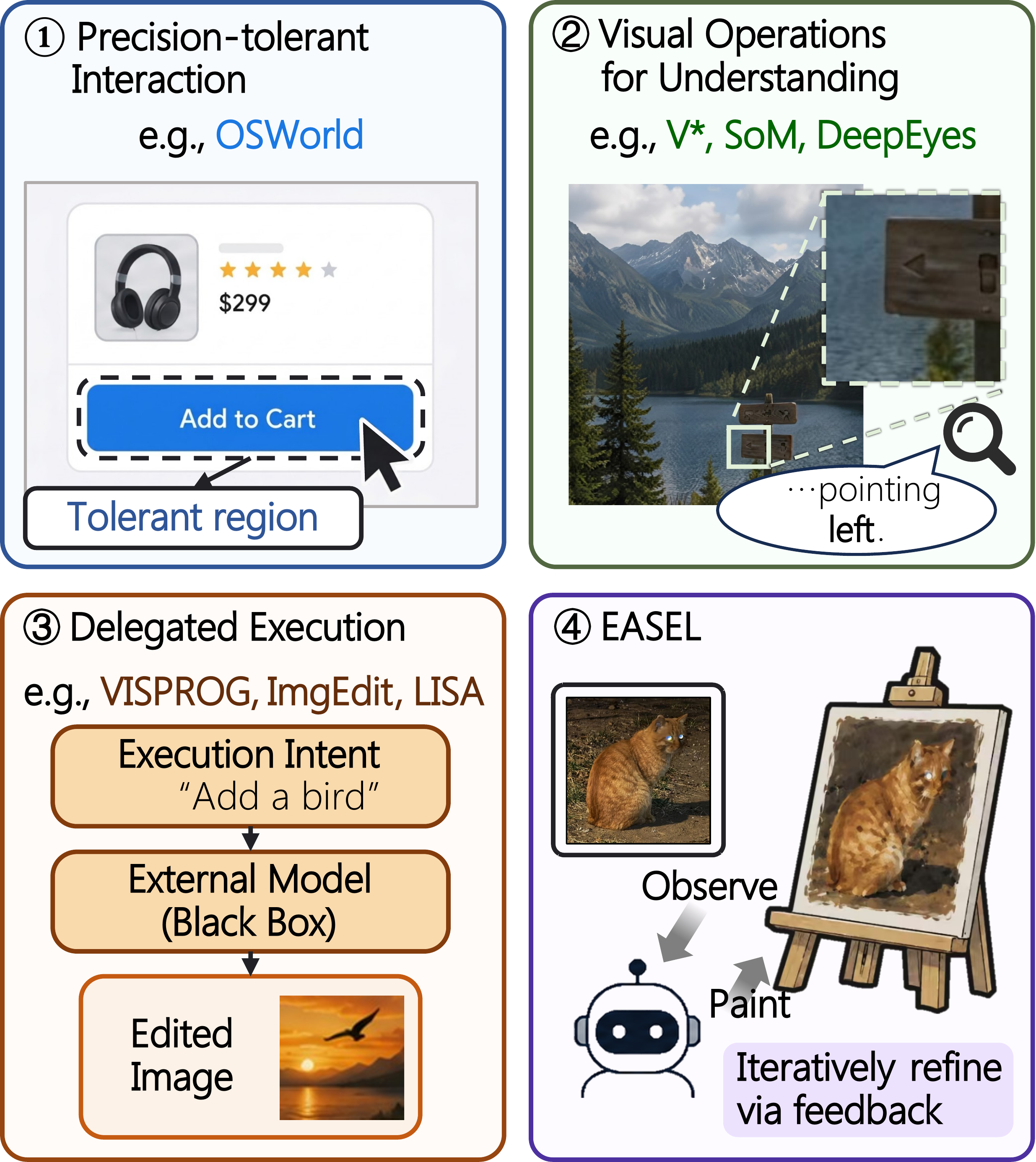}
\caption{Approaches to visual task execution. Existing paradigms each avoid precise execution---through tolerance margins, understanding-only operation, or delegation to external models. EASEL requires the agent to directly produce parameterized actions through closed-loop refinement.
}
\label{fig:approaches}
\end{figure}

Multimodal models increasingly complete tasks through external tools in complex digital environments. Evaluation has followed suit, expanding beyond answer-only QA toward tool-mediated execution in software engineering \cite{SWE-bench}, web navigation \cite{Mind2Web, WebArena}, desktop GUI use \cite{OSWorld}, and API calling \cite{BFCL}. Even in the most visual of these settings, however, interaction remains \textbf{precision-tolerant}: actions succeed as long as they fall within a generous bounding region, so precise spatial control is never strictly demanded.

Beyond such structured interfaces, recent efforts have extended agent capabilities into open-ended visual domains through visual search \cite{Vstar, ZoomEye, Zoom-Refine}, region annotation \cite{SoM, Ferret, Shikra}, pixel-level segmentation via specialized decoders \cite{LISA, PSALM, SAM4MLLM}, code-driven visual programs \cite{VISPROG, ViperGPT}, and generative editing \cite{MagicBrush, EmuEdit, ImgEdit}. As Figure~\ref{fig:approaches} illustrates, these approaches either aid understanding without producing persistent visual output, or delegate execution to external segmentation and generative models. Neither paradigm requires the agent to precisely control action parameters that directly form the result. To genuinely manipulate visual content, agents must go beyond high-level delegation to perform fine-grained spatial actions, which demands translating fine-grained visual understanding directly into precise execution parameters.

By analogy with dexterous manipulation in robotics, where the manipulator's own motor actions directly shape physical outcomes through continuous feedback rather than issuing commands to an external actuator, we term this capability \textbf{dexterous visual tool use}. 
We operationalize it as \emph{fine-grained, closed-loop parameterized visual action}:
at each step, the model reads pixel-level visual cues to infer explicit tool parameters,
those parameters directly govern the visual output, and the updated state
drives subsequent refinement.
What makes this demanding is not per-step perfection, but the need to continuously read fine-grained visual signals and translate them into corrective parameterized actions, without delegating to an external executor.

Parameterized visual action is achievable in restricted settings: neural painting \cite{LearningtoPaint, PaintTransformer} and SVG generation \cite{DiffVG} via task-specific architectures, and SimpleSeg \cite{SimpleSeg} via direct polygon prediction in general-purpose models. SimpleSeg demonstrates that highly fine-grained, even pixel-level perception is achievable natively within general-purpose MLLMs---but in a one-shot, static setting. Whether such perception can sustain across multi-step closed-loop execution, where each action changes the visual state and demands fresh spatial reasoning, remains untested; current evaluations center on task completion, where tolerance margins, abstraction, or external models absorb imprecision.

Painting offers a natural proxy for this capability: each stroke visibly changes the canvas and shapes the next decision, while parameter errors are not absorbed by a tolerant interface or external generator. Evaluation instances moreover require no manual annotation: reference images can be sourced freely, and stroke trajectories are derived automatically, making the setting easy to scale.

Guided by these properties, we introduce \textbf{EASEL}, whose primary task is reference-guided visual reconstruction: the agent observes a reference image and the current canvas, then issues incremental drawing actions, exposing the gap between semantic understanding and geometric precision. EASEL further probes the same parameterized-action interface through annotation, handwriting, and path-planning tasks driven by natural-language instructions; its annotation setting offers, to our knowledge, the first closed-loop evaluation in which general-purpose agents perform segmentation-style action without dedicated segmentation models. To examine whether trajectory-level supervision can improve this capability, we additionally construct EASEL-Data and train EASEL-9B via a two-stage curriculum: first on early-phase strokes, then extending supervision to mid-to-late steps.

\begin{enumerate}
\item We propose \textbf{EASEL}, operationalizing dexterous visual tool use as fine-grained, closed-loop parameterized visual action through 110 reference-guided reconstruction samples and 5 instruction-conditioned semantic tasks, covering up to 11{,}392 interactions per full evaluation.
\item We construct \textbf{EASEL-Data} via a scalable two-stage curriculum pipeline that converts stroke-based rendering trajectories into 440k next-action supervision samples (C1: early reconstruction; C2: light completion), and train \textbf{EASEL-9B} to examine the effect of trajectory supervision on closed-loop visual action.
\item We systematically evaluate 25 representative multimodal agents across both result quality and trajectory quality, revealing two distinct failure modes in closed-loop visual action---early saturation and post-peak degradation---and sharp capability boundaries on semantic tasks invisible to reconstruction metrics alone.
\end{enumerate}

\begin{figure*}[t]
\centering
\includegraphics[width=0.95\linewidth]{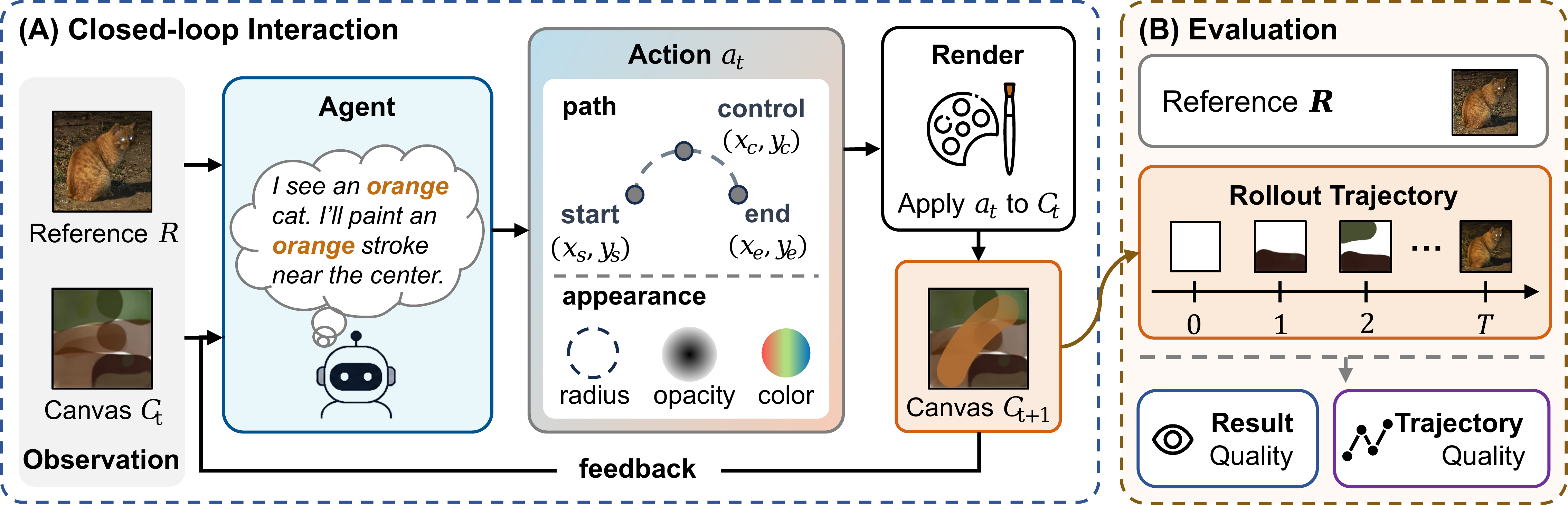}
\caption{\textbf{EASEL overview.} \textbf{(A)} At each step, the agent observes reference $R$ and current canvas $C_t$, outputs a parameterized \texttt{brush\_stroke} action, and receives the updated canvas as feedback. \textbf{(B)} The full rollout is evaluated along two axes: Result Quality and Trajectory Quality.}
\label{fig:overview}
\end{figure*}

\section{Related Work}

\subsection{Agent Evaluation and Visual Tool Use}
Agent benchmarks evaluate goal-directed execution in interactive environments, covering software engineering \cite{SWE-bench}, web and GUI navigation \cite{Mind2Web, WebArena, OSWorld}, and long-horizon visual tool use \cite{AgentVista}. GUI grounding work such as SeeClick and ScreenSpot-Pro studies screen coordinate prediction \cite{SeeClick, ScreenSpot-Pro}, but coordinates typically locate high-tolerance interactive elements and their precision does not directly determine result quality.

A separate line of work lets models operate directly on images for understanding and reasoning: through active visual search and region refinement \cite{Vstar, ZoomEye, DeepEyes, Zoom-Refine}, region annotation and grounding \cite{SoM, Ferret}, drawable workspaces \cite{VisualSketchpad, MVoT}, and program or tool invocation \cite{VISPROG, ViperGPT, Thyme, DeepEyesV2}. Segmentation-capable MLLMs such as LISA, PSALM, and SAM4MLLM further extend to pixel-level annotation through mask decoders \cite{LISA, PSALM, SAM4MLLM}, where segmentation is produced by a specialized interface rather than the agent's own actions. SimpleSeg instead predicts polygon coordinate sequences directly \cite{SimpleSeg}, instantiating parameterized visual action without decoder delegation---the closest existing paradigm to dexterous visual tool use, though it operates one-shot without closed-loop refinement.

In visual editing and production, instruction-based image editing modifies images from text instructions \cite{MagicBrush, EmuEdit, Step1X-Edit, ImgEdit}, and tool-augmented design benchmarks connect VLMs to real design software \cite{CANVAS}. These works extend agent capability from ``look and answer'' to ``use visual tools to complete tasks.'' EASEL shares this trend but differs in that parameter precision directly determines result quality.

\subsection{Parameterized Drawing and Generation}
Neural painting and stroke-based rendering are closest to EASEL in task form.
Learning to Paint and Paint Transformer decompose target images into stroke sequences
via RL and feed-forward prediction respectively \cite{LearningtoPaint, PaintTransformer}.
These works optimize task-specific models for reconstruction quality, whereas EASEL
uses reconstruction as a diagnostic lens to evaluate general-purpose agents under a
unified tool interface.
Draw with Thought reconstructs scientific diagrams as structured code \cite{DrawwithThought};
SketchAgent drives sketch creation from language descriptions \cite{SketchAgent}.
Neither targets closed-loop refinement toward an explicit visual goal, which is
EASEL's distinguishing focus.

\section{EASEL}
\subsection{Task Formulation}
EASEL centers on a \emph{reconstruction} task; semantic tasks extend the same
closed-loop framework to instruction-guided drawing (Figure~\ref{fig:overview}).

\paragraph{Reconstruction.}
Given a reference $R$ and blank canvas $C_0$, at each step $t$ the agent observes $(R, C_t)$ and outputs a \texttt{brush\_stroke} action $a_t$ in JSON;
the renderer updates $C_{t+1} = \text{Render}(C_t, a_t)$.
The episode runs for $T$ steps with the goal of making $C_T$ as visually close to $R$ as possible.

\paragraph{Semantic tasks.}
The agent observes a task image $R$ and a text instruction $l$, aiming to produce a drawing that satisfies the instruction $l$ (region outline, handwriting, or maze path).
The action space extends to $\{\texttt{brush\_stroke}, \texttt{undo}, \texttt{submit}\}$.
The brush appearance is fixed (fully opaque dark-blue, medium width), and the model predicts only the path parameters.
In maze tasks, wall-colliding actions are rejected as no-ops but consume one step, with collision feedback passed to subsequent context.

\paragraph{Model interface.}
The default interface is minimal: in reconstruction, the model observes the reference image, the current canvas, and the brush tool schema.
Semantic tasks additionally preserve a 3-turn history window.

\subsection{Benchmark Construction}

\begin{table}[t]
\centering
\small
\begin{tabular}{lccc}
\toprule
Category & \#Samples & Resolution & Budget \\
\midrule
Programmatic & 15 & 64$\times$64 & 32 \\
Spatial Geometry & 40 & 128$\times$128 & 96 \\
Abstract/Cartoon & 15 & 128$\times$128 & 96 \\
Natural Images & 40 & 128$\times$128 & 128 \\
\midrule
Semantic Tasks & 5 & 64--512 & 64--128 \\
\bottomrule
\end{tabular}
\caption{EASEL benchmark categories.}
\label{tab:task-categories}
\end{table}

EASEL includes 110 reconstruction samples and 5 semantic tasks across five categories (Table~\ref{tab:task-categories}).

The four reconstruction categories form a difficulty gradient along geometric structure, color distribution, and semantic prior strength. Programmatic provides the cleanest baseline for tool format and spatial parameterization; Spatial Geometry tests multi-object layout; Abstract/Cartoon weakens semantic priors, forcing reliance on visual evidence; Natural Images is the primary source of difficulty variation. Semantic Tasks are reported separately and include two region-annotation tasks (circle outline and left-apple outline), one handwriting task (write ``hello''), and two constrained path-planning tasks (simple maze and harder maze path). For annotation and maze tasks, the initial canvas is the task image itself and the model draws over it; for handwriting, the model writes on a blank canvas. Programmatic, maze, and Circle Outline tasks are procedurally synthesized; remaining categories are generated via GPT-image-2.

\subsection{Evaluation Metrics}
\label{sec:metrics}

\begin{figure*}[t]
  \includegraphics[width=\linewidth]{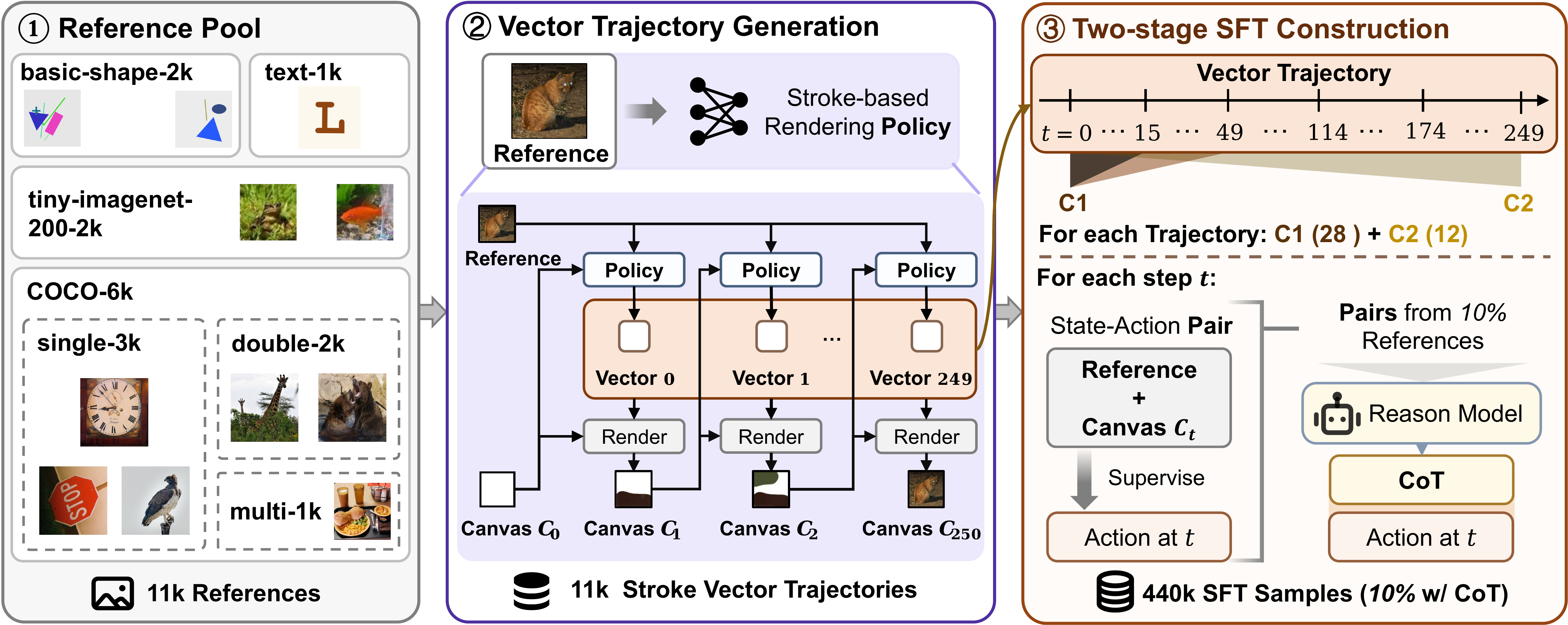}
  \caption{
  EASEL-Data construction pipeline. 
  \textbf{(1) Reference Pool:} 11k reference images spanning four source domains. \textbf{(2) Vector Trajectory Generation:} a stroke-based rendering policy produces a 250-step vector trajectory per reference. \textbf{(3) Two-stage SFT Construction:} C1 densely covers early strokes (0--49, with full retention through stroke 16); C2 spans the full trajectory (0--249), yielding 440k next-action prediction samples. For 10\% of references, a reasoning model annotates all corresponding trajectory samples with CoT supervision (44k CoT samples).
  }
\label{fig:data}
\end{figure*}

\paragraph{Overview.}
EASEL evaluates each rollout along two axes. \textbf{Result Quality} measures whether the final visual output satisfies the target; \textbf{Trajectory Quality} diagnoses how the model uses intermediate feedback over the action sequence.

\paragraph{Reconstruction.} We compute SSIM, normalized RGB $L_1$ distance, and Edge IoU, and define the similarity $S_t$ as:
\begin{align}
S_t = &\ 0.5 \cdot \text{SSIM} \notag\\
&+ 0.3 \cdot (1 - \text{RGB}_{L_1}) \notag\\
&+ 0.2 \cdot \text{Edge IoU}
\end{align}
Weights encode the intended balance: SSIM primary, RGB fidelity for appearance, Edge IoU for boundary structure. Ranking sensitivity across alternative weightings is reported in the supplementary material.

For standard reconstruction, Result Quality is measured by Final Similarity $S_T$, the primary ranking metric. Trajectory Quality is measured by four diagnostics: Similarity@50\% ($S_{\lfloor T/2 \rfloor}$), Trajectory AUC ($\frac{1}{T}\sum_t S_t$), Best Similarity ($\max_t S_t$), and Final-Best Gap ($\max_t S_t - S_T$). F-B Gap diagnoses retention of the best achieved state and should be interpreted jointly with Final Similarity: a small gap can indicate either a strong result that was preserved or limited progress with little post-peak loss.

\paragraph{Semantic Tasks.} For semantic tasks, Result Quality is the task-specific score $\in [0, 1]$: (1) outline tasks (Circle Outline, Apple Outline) use the Dice coefficient between the model's drawn region and the target mask; (2) handwriting (Hello) uses a multimodal LLM to recognize each letter and compute positional match rate; (3) maze tasks (Simple Maze, Maze) compute the normalized progress of the model's path along the reference route. Semantic trajectories also log action traces, invalid action rate, \texttt{undo}/\texttt{submit} behavior, and intermediate canvases for failure analysis. Full semantic task specifications are in the supplementary material.

\subsection{Tool Semantics}
\label{sec:tool-semantics}

The \texttt{brush\_stroke} action comprises 13 continuous normalized parameters:
\begin{equation}
\begin{aligned}
a ={}& (
\underbrace{x_s, y_s, x_c, y_c, x_e, y_e}_{\text{path}}, \\
& \underbrace{r_s, \alpha_s, r_e, \alpha_e}_{\text{radius \& opacity}},
  \underbrace{c_r, c_g, c_b}_{\text{color}}
)
\end{aligned}
\end{equation}

The path parameters $(x_s, y_s)$, $(x_c, y_c)$, $(x_e, y_e)$ define a quadratic Bézier curve in normalized canvas coordinates. Appearance parameters include per-endpoint radius and opacity $(r_s, \alpha_s, r_e, \alpha_e)$, linearly interpolated along the stroke, and a shared RGB color $(c_r, c_g, c_b)$. 
A quadratic Bézier primitive keeps the schema compact while exposing the continuous path and appearance parameters that constitute the target capability. 
Semantic tasks provide a reduced-schema setting by fixing appearance and requiring only the six path coordinates. The stroke is composited onto the canvas via alpha blending. The full schema and renderer specification are in the supplementary material.

\section{EASEL-Data}

EASEL-Data construction proceeds in three stages: reference image preparation, vector trajectory generation, and two-stage SFT construction (Figure~\ref{fig:data}).

\paragraph{Reference pool.} We prepare 11k reference images from four sources: procedurally generated geometry and text/glyphs, single-/dual-/multi-subject crops from COCO \cite{COCO}, and natural images from Tiny ImageNet \cite{tiny}, spanning a visual complexity spectrum from simple geometry to complex natural scenes. Detailed category specifications are in the supplementary material.

\paragraph{Vector trajectory pool.}
For each reference image, we generate a 250-step \texttt{brush\_stroke} trajectory
using a stroke-based painting policy (\emph{data policy}) \cite{LearningtoPaint}.
This 250-step run is the \emph{data-native} setting; a \emph{budget-aligned} variant
($\lceil B/5 \rceil$ policy steps, five raster strokes each) serves as a score anchor
in Section~\ref{sec:experiments}.

\paragraph{Two-stage SFT construction.} We convert trajectories into next-action prediction: given the reference and current canvas, the model predicts the next stroke, with canvases produced by replaying prior actions. Each supervision target is one observed continuation from the rendering policy.

We construct a two-stage curriculum: \textbf{C1} (Early Reconstruction, 308k samples) retains all steps from strokes 0--15 and samples from strokes 16--49, ensuring every blank-canvas state is covered and training the model to establish coarse structure and color; \textbf{C2} (Light Completion, 132k samples) spans the full trajectory with decreasing density to cover mid-to-late refinement. Spatial coordinates are quantized to integers in $[0,1000]$ to align with MLLM tokenization. A reasoning model additionally generates CoT supervision for 10\% of references, yielding 44k CoT samples. Full construction specifications are in the supplementary material.

\subsection{EASEL-9B}

We fine-tune Qwen3.5-9B \cite{QWEN35} in two stages using LoRA (rank 64, alpha 128, all-linear targets). Stage 1 trains on C1 with learning rate $1\times10^{-4}$; Stage 2 continues from Stage 1 adapters on C2 with 15\% C1 replay at a reduced learning rate $5\times10^{-5}$. Both stages use bf16 precision, DeepSpeed ZeRO-2, effective batch size 64 (8 GPUs), max sequence length 8192, AdamW with weight decay 0.1 and cosine schedule. The ViT last four blocks and merger layers receive full gradient updates via \texttt{modules\_to\_save}. EASEL-9B is not intended as an upper bound on painting quality, but as a diagnostic probe for the effect of the two-stage curriculum. Full training hyperparameters are in the supplementary material.

\begin{table}[t]
\centering
\small
\begin{tabular}{lcc}
\toprule
Control / Agent & Final Sim.$\uparrow$ & Edge IoU$\uparrow$ \\
\midrule
Blank-white canvas & 0.446 & 0.000 \\
Reference mean-color fill & 0.534 & 0.000 \\
Reference dominant-color fill & 0.531 & 0.000 \\
Gemini 3.1 Pro (best MLLM) & 0.535 & 0.053 \\
\midrule
Data policy, budget-aligned  & 0.665 & 0.281 \\
Data policy, native (250-step) & 0.715 & 0.563 \\
\bottomrule
\end{tabular}
\caption{Reconstruction score anchors. Uniform fills score zero on Edge IoU; the data policy provides a reference point for attainable scores under two step-budget settings.}
\label{tab:reconstruction-controls}
\end{table}

\begin{table*}[t]
\centering
\setlength{\tabcolsep}{6pt}
\small
\begin{tabular}{lcccc|c|cccc}
\toprule
\multicolumn{1}{c}{Model} & \multicolumn{4}{c|}{Category Breakdown} & \multicolumn{1}{c|}{Overall} & \multicolumn{4}{c}{Trajectory} \\
\cmidrule(lr){2-5} \cmidrule(lr){6-6} \cmidrule(l){7-10}
Model & Prog. & Spatial & Abs. & Nat. & Final$\uparrow$ & @50\%$\uparrow$ & AUC$\uparrow$ & Best$\uparrow$ & F-B Gap$\downarrow$ \\
\midrule
Gemini 3.1 Pro & \textbf{0.598} & \textbf{0.579} & \textbf{0.521} & \textbf{0.472} & \textbf{0.535} & \textbf{0.532} & \textbf{0.534} & \textbf{0.566} & 0.032 \\
Claude Opus 4.7 & 0.469 & \underline{0.531} & 0.468 & \underline{0.415} & \underline{0.472} & \underline{0.473} & \underline{0.474} & \underline{0.506} & 0.035 \\
Claude Sonnet 4.6 & 0.439 & 0.496 & 0.475 & 0.390 & 0.447 & 0.450 & 0.451 & 0.478 & 0.031 \\
GLM-5V-Turbo & 0.367 & 0.495 & 0.430 & 0.382 & 0.428 & 0.431 & 0.433 & 0.487 & 0.060 \\
GPT-5.5 & 0.434 & 0.467 & 0.430 & 0.380 & 0.426 & 0.436 & 0.440 & 0.499 & 0.073 \\
qwen3.5-plus & 0.412 & 0.482 & 0.464 & 0.367 & 0.428 & 0.433 & 0.434 & 0.460 & 0.032 \\
\midrule
Qwen3.6-27B & 0.489 & 0.476 & 0.498 & 0.364 & 0.440 & 0.440 & 0.441 & 0.454 & 0.014 \\
Qwen3.5-35B-A3B & 0.483 & 0.490 & 0.495 & 0.359 & 0.442 & 0.442 & 0.442 & 0.451 & \underline{0.009} \\
Qwen3.5-27B & 0.469 & 0.473 & 0.486 & 0.361 & 0.433 & 0.433 & 0.434 & 0.451 & 0.018 \\
Qwen3.5-9B & 0.473 & 0.474 & 0.475 & 0.359 & 0.432 & 0.432 & 0.433 & 0.447 & 0.015 \\
Qwen3.5-4B & 0.482 & 0.484 & 0.486 & 0.362 & 0.440 & 0.440 & 0.420 & 0.448 & \textbf{0.008} \\
Qwen3-VL-32B-Thinking & 0.482 & 0.481 & 0.490 & 0.360 & 0.438 & 0.438 & 0.416 & 0.447 & \textbf{0.008} \\
Qwen3-VL-30B-A3B-Inst. & 0.421 & 0.491 & 0.469 & 0.361 & 0.431 & 0.433 & 0.433 & 0.464 & 0.032 \\
Qwen3-VL-30B-A3B-Think. & 0.416 & 0.478 & 0.484 & 0.355 & 0.426 & 0.428 & 0.410 & 0.455 & 0.029 \\
kimi-k2.5 & 0.459 & 0.490 & 0.481 & 0.374 & 0.442 & 0.444 & 0.445 & 0.464 & 0.022 \\
InternVL3-14B & 0.472 & 0.438 & 0.448 & 0.344 & 0.420 & 0.420 & 0.420 & 0.451 & 0.019 \\
InternVL3-8B & 0.483 & 0.448 & 0.466 & 0.346 & 0.418 & 0.418 & 0.420 & 0.447 & 0.029 \\
Llama 4 Maverick & 0.480 & 0.475 & 0.490 & 0.360 & 0.437 & 0.437 & 0.437 & 0.445 & \textbf{0.008} \\
Llama 4 Scout & 0.473 & 0.470 & 0.480 & 0.355 & 0.430 & 0.430 & 0.430 & 0.438 & \textbf{0.008} \\
Llama 3.2 Vision 90B & 0.460 & 0.455 & 0.465 & 0.348 & 0.422 & 0.422 & 0.422 & 0.435 & 0.013 \\
Llama 3.2 Vision 11B & 0.455 & 0.450 & 0.460 & 0.345 & 0.415 & 0.415 & 0.415 & 0.428 & 0.013 \\
Gemma3-27B & 0.470 & 0.465 & 0.472 & 0.352 & 0.426 & 0.426 & 0.426 & 0.440 & 0.014 \\
DeepSeek-VL2 & 0.452 & 0.445 & 0.455 & 0.342 & 0.414 & 0.414 & 0.414 & 0.428 & 0.014 \\
MiMo-VL-7B-RL & 0.411 & 0.442 & 0.444 & 0.348 & 0.404 & 0.406 & 0.409 & 0.451 & 0.047 \\
\midrule
EASEL-9B (w/o C2) & \underline{0.505} & 0.491 & 0.486 & 0.391 & 0.456 & 0.455 & 0.456 & 0.475 & 0.020 \\
EASEL-9B & 0.495 & 0.499 & \underline{0.501} & 0.390 & 0.459 & 0.460 & 0.459 & 0.478 & 0.019 \\
\bottomrule
\end{tabular}
\caption{EASEL standard reconstruction results and trajectory diagnostics. Final is the primary ranking metric; @50\%, AUC, Best, and F-B Gap are trajectory diagnostics defined in Section~\ref{sec:metrics}. Models are grouped by family; EASEL-9B variants are listed separately as a curriculum ablation. \textbf{Bold} indicates the best result in each column; \underline{underline} indicates the second best.}
\label{tab:reconstruction}
\end{table*}

\section{Experiments}
\label{sec:experiments}

We evaluate 25 multimodal agents, including 6 closed-source APIs and 19 open-source models spanning 4B--90B parameters selected to cover the current performance frontier. The evaluated model families include proprietary Gemini, Claude, GPT, GLM, and Qwen APIs \cite{Gemini31Pro,ClaudeSystemCards,GPT55,GLM5VTurbo,QWEN35}, as well as open-source Qwen, InternVL, Llama, Gemma, DeepSeek-VL, MiMo-VL, and Kimi models \cite{QWEN35,Qwen3VL,Qwen36_27B,InternVL3,Llama4,Llama32Vision,Gemma3,DeepSeekVL2,MiMoVL,KimiK25}.

\subsection{Standard Reconstruction}

\begin{figure}[t]
\centering
\includegraphics[width=\columnwidth]{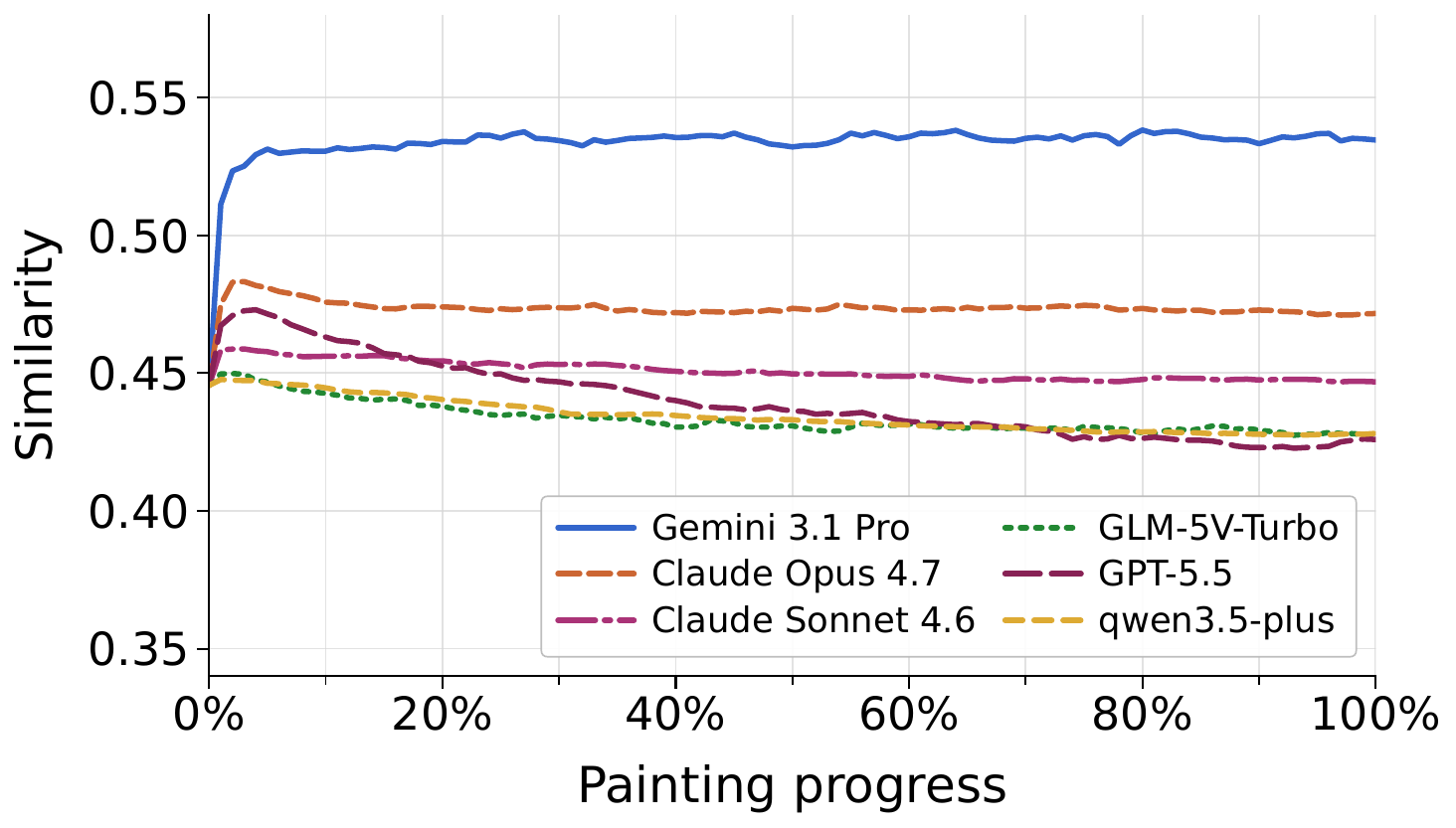}
\caption{Average similarity over painting progress across reconstruction tasks.
Most gains occur within the first 10\% of the step budget, after which trajectories plateau or degrade.}
\label{fig:trajectory}
\end{figure}

\begin{table*}[t]
\centering
\setlength{\tabcolsep}{4.5pt}
\small
\begin{tabular}{lccccc|cccc}
\toprule
\multicolumn{1}{c}{Model}
& \multicolumn{5}{c|}{Result Quality}
& \multicolumn{4}{c}{Trajectory} \\
\cmidrule(lr){2-6} \cmidrule(l){7-10}
Model
& Circle$\uparrow$
& Apple$\uparrow$
& Hello$\uparrow$
& S-Maze$\uparrow$
& H-Maze$\uparrow$
& \shortstack{Maze\\Inv.$\downarrow$}
& \shortstack{Maze\\Undo}
& \shortstack{Non-Maze\\Undo}
& \shortstack{Avg.\\Submit\\(\%)} \\
\midrule
Gemini 3.1 Pro
& \textbf{0.919} & \textbf{0.783} & \textbf{1.000} & 0.183 & 0.007
& 0.21 & 87 & 0 & 43.5 \\
GPT-5.5
& 0.785 & 0.761 & \textbf{1.000} & \textbf{1.000} & 0.031
& \textbf{0.19} & 81 & 0 & 41.6 \\
Claude Sonnet 4.6
& 0.521 & 0.491 & 0.200 & 0.157 & \textbf{0.047}
& 0.93 & 0 & 1 & 48.7 \\
Claude Opus 4.7
& 0.000 & 0.602 & 0.200 & 0.026 & 0.000
& 0.87 & 3 & 124 & 87.7 \\
GLM-5V-Turbo
& 0.484 & 0.722 & 0.000 & 0.000 & 0.000
& 1.00 & 0 & 0 & 43.0 \\
\bottomrule
\end{tabular}
\caption{Semantic task results and trajectory diagnostics. Circle and Apple are outline Dice scores; Hello is letter-level recognition accuracy; S-Maze and H-Maze are normalized path-progress scores. Maze Inv.: fraction of invalid actions on maze tasks; Maze Undo / Non-Maze Undo: undo counts on maze vs.\ outline+handwriting tasks; Avg.\ Submit\%: mean of per-episode submit percentile (submit step $\div$ episode budget $\times 100$), with budget-exhausted episodes counted as 100\%. \textbf{Bold} indicates best result per quality column and lowest Maze Inv.}
\label{tab:semantic}
\end{table*}

Table~\ref{tab:reconstruction} reports reconstruction results and trajectory diagnostics; Figure~\ref{fig:trajectory} plots the average reconstruction curves for the six closed-source models.

\paragraph{Score anchors and task solvability.}
Table~\ref{tab:reconstruction-controls} anchors the absolute score range.
The best agent, Gemini 3.1 Pro (0.535), approximately matches the reference mean-color fill (0.534) in composite similarity, although Gemini captures edge structure (Edge IoU 0.053) while all uniform fills score zero on Edge IoU.
The data policy reaches 0.665 under the budget-aligned setting (Edge IoU 0.281) and 0.715 in the data-native setting (Edge IoU 0.563).
Both substantially exceed every evaluated agent, confirming that the task is not inherently unsolvable at this score range.
The continuous span from uniform fills (Edge IoU 0) to Gemini (Edge IoU 0.053) to the data policy (Edge IoU 0.281--0.563) also confirms that the composite metric retains discriminative range well above current MLLMs, rather than saturating near 0.53.

\paragraph{Overall performance.}
Overall performance is low and clearly stratified. Gemini 3.1 Pro achieves the best Final Similarity (0.535), followed by Claude Opus 4.7 (0.472), while open-source models cluster in the 0.40--0.45 range with small inter-model variance and no clear correlation with parameter count. EASEL-9B (full curriculum) reaches 0.459, ranking third overall and the highest score among non-proprietary models, outperforming its base model Qwen3.5-9B (0.432) by +0.027. Across categories, Natural Images is hardest due to complex textures and fine-grained color distributions, whereas Programmatic is easiest due to strong geometric priors. These results show that EASEL exposes a systematic weakness in closed-loop visual action.

\paragraph{Trajectory patterns.}
Trajectory diagnostics reveal two qualitatively distinct behavioral patterns among closed-source models.
Gemini 3.1 Pro, Claude Opus 4.7, and Claude Sonnet 4.6 rise sharply within the first 5\% of steps and then plateau: their @50\%, AUC, and Final scores are nearly identical, indicating these models effectively stop making meaningful changes after early gains.
By contrast, GPT-5.5, GLM-5V-Turbo, and qwen3.5-plus continue acting but degrade: AUC exceeds Final in all three cases, with GPT-5.5 showing the strongest degradation (F-B Gap 0.073; AUC 0.440 vs.\ Final 0.426).
The distinction reflects different local action tendencies: plateau models tend toward conservative outputs once visual gain diminishes, while degrading models continue generating novel actions regardless of canvas state.
Across all closed-source models, the effective peak is reached within the first 10\% of the step budget, indicating that neither strategy exploits the remaining feedback signal.

\paragraph{F-B Gap interpretation.}
F-B Gap measures retention, not quality, and its interpretation depends on the underlying behavior.
The five open-weight models with F-B Gap below 0.01 have exact-repeat rates of 82.3\%--92.4\%, so their small gaps reflect near-total inaction rather than active quality maintenance.
Closed-source plateau models achieve comparable F-B Gaps ($\leq$0.035) through selectively conservative actions, which---unlike verbatim repetition---implies genuine sensitivity to canvas state; Final Similarity separately captures the quality of what is being preserved.
The gap between closed-source plateau models (0.47--0.54) and open-source near-inaction models (0.40--0.45) suggests stronger models read visual evidence more effectively early on. Yet even top closed-source models plateau well below the data policy's 0.665--0.715: the bottleneck is not initial perception but sustained, feedback-driven correction across the trajectory.

\paragraph{Trajectory supervision.}
We compare Qwen3.5-9B (base, 0.432) with EASEL-9B trained on C1 only (0.456, +0.024) and the full C1+C2 curriculum (0.459, +0.027 over base). C1 training alone accounts for the majority of the gain, establishing coarse structure and color from early strokes; adding C2 contributes a further +0.003, indicating that mid-to-late completion supervision provides modest but measurable improvement. Final-Best Gap increases slightly from 0.015 (base) to 0.019--0.020 (trained), suggesting that trajectory supervision improves local action quality while long-horizon feedback correction remains limited under the current setup.

\subsection{Semantic Task Analysis}

\begin{figure}[t]
\centering
\includegraphics[width=0.96\columnwidth]{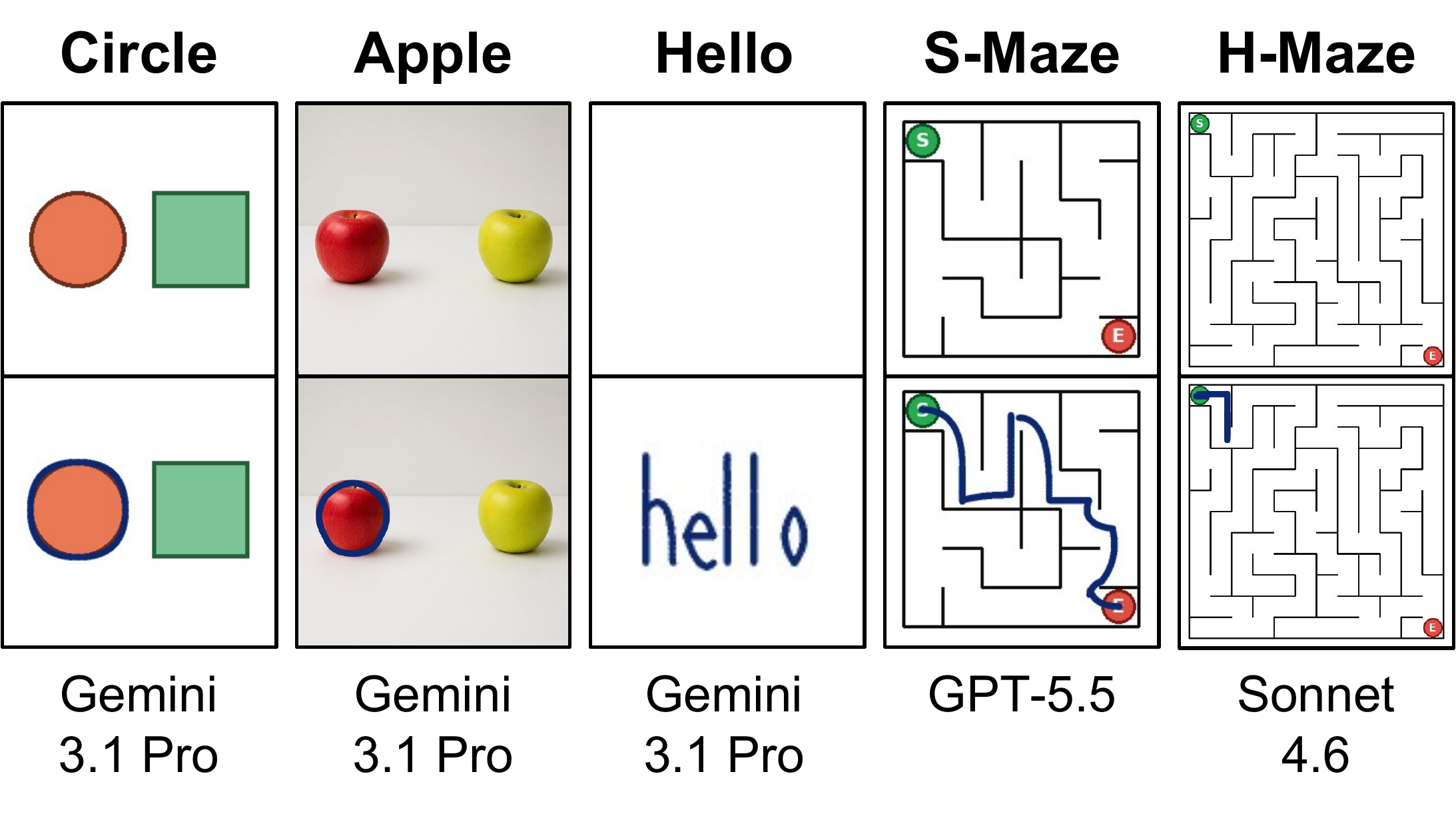}
\caption{Best results on each semantic task. \textbf{Top row}: starting canvas. \textbf{Bottom row}: model output (Gemini 3.1 Pro for Circle, Apple, and Hello; GPT-5.5 for S-Maze; Claude Sonnet 4.6 for H-Maze).}
\label{fig:semantic}
\end{figure}

We analyze five representative models with nontrivial and differentiated semantic
behavior (Table~\ref{tab:semantic}); models producing near-zero scores on all tasks
are omitted. Semantic tasks reveal sharper capability boundaries than reconstruction.

\paragraph{Outline and handwriting.}
Top closed-source models can perform segmentation-style annotation through direct stroke control: Gemini 3.1 Pro and GPT-5.5 lead both outline tasks and achieve perfect handwriting, while weaker models produce partial or zero scores.
Claude Opus 4.7 fails on Circle Outline entirely due to persistent draw-undo loops---action strategy stability is a prerequisite for tasks requiring cumulative canvas accumulation.

\paragraph{Path planning.}
GPT-5.5 achieves a perfect score on Simple Maze despite weaker reconstruction performance, confirming that reconstruction precision and instruction-following are partially orthogonal dimensions. The harder Maze Path remains unsolved across all models.

\paragraph{Action traces.}
Gemini 3.1 Pro and GPT-5.5 use \texttt{undo} almost exclusively on maze tasks, suggesting targeted collision recovery rather than general uncertainty; GPT-5.5 shows iterative correction on Simple Maze, submitting after 25 undos with a perfect score---the clearest example of closed-loop self-correction in our evaluation.
Claude Opus 4.7 inverts this pattern, with 124 non-maze undos and only 3 maze undos, consistent with unstable draw-undo loops that erase useful strokes and prevent accumulation.
Claude Sonnet 4.6 almost never uses \texttt{undo} yet produces near-maximal invalid rates on maze tasks, indicating it neither detects nor responds to constraint violations---a different failure mode in which the model proceeds confidently despite errors.

\section{Discussion}

\paragraph{Scope and proxy validity.} EASEL trades breadth for control: the 2D canvas isolates visual goals, parameterized actions, and closed-loop feedback from confounding variables, enabling focused diagnosis of fine-grained visual manipulation and closed-loop correction under a controllable protocol.

\paragraph{Applications.}
In our setting, agents act on visual content through explicit, editable strokes rather than delegating execution to specialized modules or generative systems.
The semantic tasks demonstrate breadth within this interface through annotation, handwriting, and path planning.
Medical annotation, CAD sketching, diagram authoring, and embodied manipulation motivate the broader capability because they share visual state observation, structured action prediction, and potential closed-loop feedback.
Beyond evaluation, the explicit action traces and environment feedback make EASEL compatible with outcome-based optimization, reinforcement learning, and planning methods.

\paragraph{Generalization of trajectory supervision.}
EASEL-9B preserves visual perception after training: on standard perception benchmarks it matches or modestly improves upon the base model---MMVP +0.7\%, POPE +2.0\%, HallusionBench +0.6\% \cite{MMVP,POPE,Hallusion}, and on BLINK's spatial reasoning and counting subtasks +0\% and +1.7\% respectively \cite{BLINK}---suggesting that closed-loop action training preserves and modestly broadens fine-grained multimodal perception. This raises the possibility that EASEL-style tasks, combined with more targeted supervision design, could serve as a training signal for fine-grained visual perception.

\section{Conclusion}

We present EASEL, a benchmark that operationalizes dexterous visual tool use as precision-critical, closed-loop parameterized visual action. By combining reference-guided reconstruction with instruction-conditioned semantic tasks and trajectory diagnostics, EASEL exposes failures obscured by answer-only QA and high-level tool-use benchmarks: models plateau early, degrade after initial gains, and rarely exploit corrective feedback. A two-stage curriculum (EASEL-9B) improves over the base model by a relative 6.3\%, with early-phase training accounting for most of the gain and fine-grained visual perception improved, while long-horizon feedback correction remains limited.

\bibliography{easel}

\end{document}